\documentclass[journal]{IEEEtran}

\usepackage{xcolor,soul,framed} %,caption

\colorlet{shadecolor}{yellow}
\usepackage[pdftex]{graphicx}
\graphicspath{{../pdf/}{../jpeg/}}
\DeclareGraphicsExtensions{.pdf,.jpeg,.png}

\usepackage[cmex10]{amsmath}
\usepackage{mathrsfs}
\usepackage{array}
\usepackage{mdwmath}
\usepackage{mdwtab}
\usepackage{eqparbox}
\usepackage{url}

\usepackage{float}
\usepackage{stfloats}

\usepackage{etoolbox}
\usepackage{latexsym}
\makeatletter
\patchcmd{\@makecaption}
  {\scshape}
  {}
  {}
  {}
\makeatletter
\patchcmd{\@makecaption}
  {\\}
  {.\ }
  {}
  {}
\makeatother

\begin{document}
\bstctlcite{IEEEexample:BSTcontrol}
    \title{Design, Modelling and Control of SPIROS:\\ The Six Propellers and Intermeshing Rotors Based Omnidirectional Spherical Robot}
  
    \author{YOGESH~PHALAK, Student Member IEEE % <-this % stops a space

\thanks{Yogesh Phalak is an Undergraduate Student at Electrical Engineering Department, Visvesvaraya National Institute of Technology, Nagpur, India.(e-mail: yogeshphalak31415@students.vnit.ac.in).}}  

% ======================================================
\maketitle

%=======================================================
%  ABSTRACT
% ======================================================

\begin{abstract}
%\boldmath

Since the past few decades, several designs and control methods have been developed for the Spherical Robots (SRs) with thoroughly analyzed mechanics on generalized 3D terrains. But the vertical motion and the steep inclination maneuver has been an unsolved problem with the existing SR's driving mechanisms. Also, the possibilities of wind-powered or air-propelled SRs have not been fully explored. This paper introduces the new Omnidirectional Spherical Robot mechanism named SPIROS: The Six Propeller and Intermeshed Rotor based Omnidirectional Spherical Robot. The SPIROS is driven by a novel octahedral arrangement of six intermeshed rotary air thrusters placed in the Goldberg Polyhedral shaped spherical gridshell. The advantage of the proposed design lies in its air-powered propulsion, which improves on the obstacle avoidance and the slope climbing abilities. The robot's dynamic models are derived using the existing kinematical models of the Continuous Rolling Spherical Robots (CR-SR), with subcategories, triple axes rolling (3R-SR), dual axes rolling (2R-SR) and rolling and turning (RT-SR) spherical robots. The path tracking control scheme based on the pure pursuit algorithm is presented. Simulations are carried out in MATLAB and Simulink to validate the developed models and the effectiveness of the proposed control schemes.

\end{abstract}

%=======================================================
%  KEYWORDS 
%=======================================================

\begin{IEEEkeywords}
Spherical Robot, Omnidirectional Robot, Intermashing Rotors, Propellors, Modelling, Mechanical Design, Controls, Mathematical Model.
\end{IEEEkeywords}

\IEEEpeerreviewmaketitle

% ======================================================
% INTRODUCTION
% ======================================================

\section{Introduction}

\IEEEPARstart{T}{he} Omnidirectional Spherical Robots are the class of mobile robots which are generally having Spherical shell and internal driving components that provide torques required for their rolling motion. Due to a perfectly symmetric geometric structure, SRs acquire multiple advantages over other mobile robots, such as improved mobility, maneuverability, omnidirectionality and the compact structure. The literature survey \cite{crossley2006literature} illustrated the different designs of SRs. Since the first contact wheeled driven SR was developed by Halme et al. in 1996 \cite{halme1996spherical}, the improved two contact wheeled configuration was given by K. Husoy \cite{Husoy2003instrumentation}. A. Bicchi et al. proposed a small resting car-propelled design of SR \cite{bicchi1997introducing}. S. Bhattacharya developed a design that involved a set of two mutually perpendicular rotors attached to the inside of the sphere \cite{bhattacharya2000design}. Joshi and Banavar used four rotors to achieve omnidirectionality in SRs \cite{joshi2010design}. While, Gajamohan et al. \cite{gajamohan2013cubli} used three rotors in a cube robot to rotate and balance omnidirectionally. SR called Rotundus produces pendulum based propulsion \cite{Rotundus}. The Spherobot developed by R. Mukherjee et al. used fixed radially distributed weights along spokes for the actuation \cite{mukherjee1999simple}. The four axle mounted tetrahedral stepper motor driven design was proposed by A. Javadi and P. Mojabi \cite{mojabi2002introducing}, Sang et al. \cite{sang2010modeling} and Tomik et al. \cite{tomik2012design}. Sugiyama et al. proposed a shape-memory alloy to deform and roll wheels and spheres \cite{sugiyama2005circular}, while Wait et al. \cite{wait2010self} and Artusi et al. \cite{artusi2011electroactive}  deform panels on a robot’s shell via air bladders and dielectric actuators.

\begin{figure}[t]
  \begin{center}
  \includegraphics[width=0.70\linewidth]{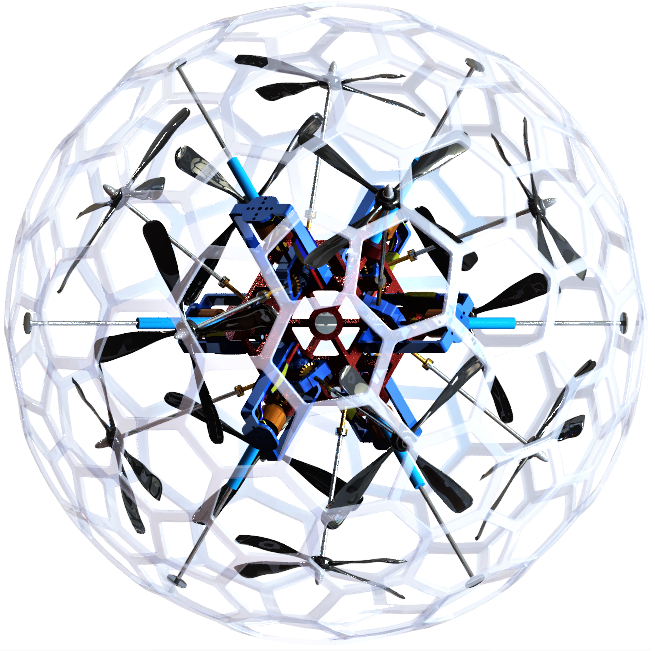}
  \caption{CAD model of the SPIROS, the proposed Omnidirectional Spherical Robot.}\label{SpirosCAD}
  \end{center}
\end{figure}

Recent researches have solved the complexity of the dynamics and control problems of the SRs and found the suitable applications of SRs in surveillance \cite{surveillance}, environmental monitoring \cite{monitoring}, exploration operations in unknown environments \cite{uenvironment}, agriculture \cite{agriculture}. Nevertheless, the vertical motion and the steep inclination maneuver has been an unsolved problem with the existing contact wheeled based designs. Also, the contact wheel's skidding, tipping over, falling or friction with the surface makes the SRs vulnerable or inefficient \cite{friction}. The deformable robot designs does not meet the level of traditional spherical rolling robots, they are somewhat limited by their tethered operation and high voltage requirements. As a result, the fully efficient functionality of traditional SR characteristics cannot be achieved. The design concept to have the sphere split apart to deploy a leg mechanism can't actuate the robot on the vertical surfaces.

\begin{figure}[h]
  \begin{center}
  \includegraphics[width=0.9\linewidth]{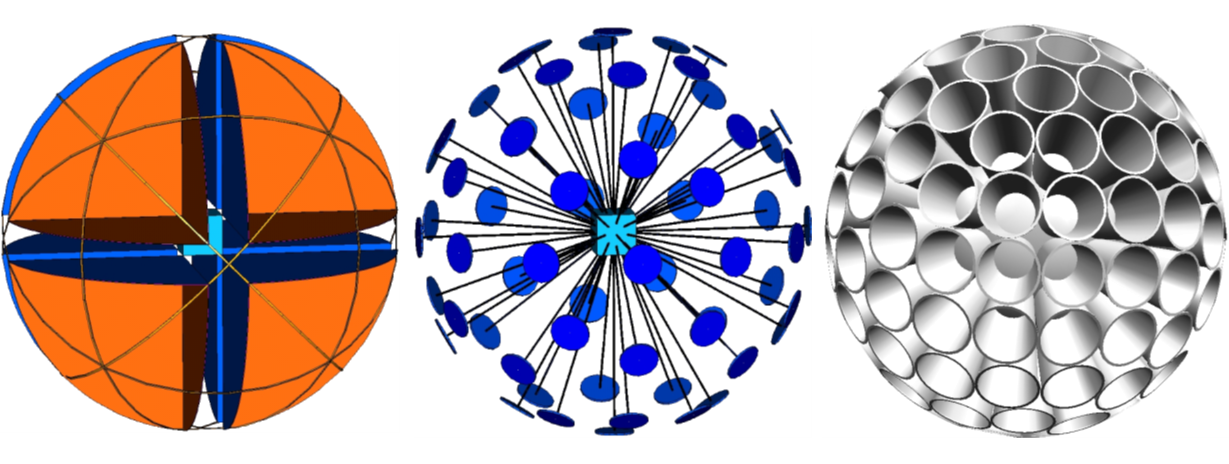}
  \caption{Tumbleweeds inspired wind-powered Wind Spheres}\label{wind_spheres}
  \end{center}
\end{figure}

Due to NASA’s efforts to create low-cost Mars exploration rovers, there has been some work on wind-powered SRs based on the biomimetic design which is inspired by tumbleweeds \cite{antol2003low} shown in Fig. \ref{wind_spheres}. Such a design would make the robot to go over the larger obstacles than a car-like rover of its size. Other designs have mixed the pendulum design with wind-powered robots, allowing them to steer and stop by using a pendulum that can be retracted into the center of the sphere to “coast” \cite{jones1998beach}. Since then the possibilities of wind-powered SRs have not been fully explored. Air-based propulsion would allow the robot to maneuver over steep as well as vertical surfaces. Moreover, by introducing the air propulsion based drive in SRs, there will be no frictional losses between the driving mechanism and the outer shell, while maintaining a smaller size than the wind spheres, the improved functionality can be achieved.

This paper presents a novel design of SR called SPIROS: The Six Propeller and Intermeshed Rotor based Omnidirectional Spherical Robot. SPIROS uses six intermeshed rotary air thrusters \cite{wildhaber1974rotor} placed in the octahedral arrangement for propulsion. It is omnidirectional (which is able to roll in any direction instantaneously) unlike most of the existing SR designs. The Goldberg Polyhedral \cite{hart2013goldberg} shaped gridshell is used to achieve the outer geometric spherical shape and to ensure the free airflow needed for the thrusters. We have described the mechanical design of the intermeshed rotary thrusters, the outer gridshell and the full robot assembly in detail. Also, as per the proposed classification and the constraints of SRs in \cite{moazami2019kinematics}, the dynamics equations for SRs rolling over 3D terrains are derived. A modified pure pursuit method \cite{corke2017robotics} is utilized for the path tracking control problem of derived models.

The remainder of this paper is organized as follows: Section II elaborates on the working principle of the robot. Section III is dedicated towards the mechanical and electrical design specifications of the robot assembly. In Section IV, the dynamic equations of SPIROS are derived based on the Continuous Rolling Spherical Robots (CR-SR) model, with triple axes rolling (3R-SR), dual axes rolling (2R-SR), and rolling and turning (RT-SR) spherical robots as subcategories. In Section V, a 3D path tracking control scheme has been presented. The simulation results in MATLAB and Simulink are presented in Section VI, followed by the conclusion and future work.

% ======================================================
% WORKING PRINCIPLE
% ======================================================

\section{Working Principle}

\begin{figure}[h]
  \begin{center}
  \includegraphics[width=\linewidth]{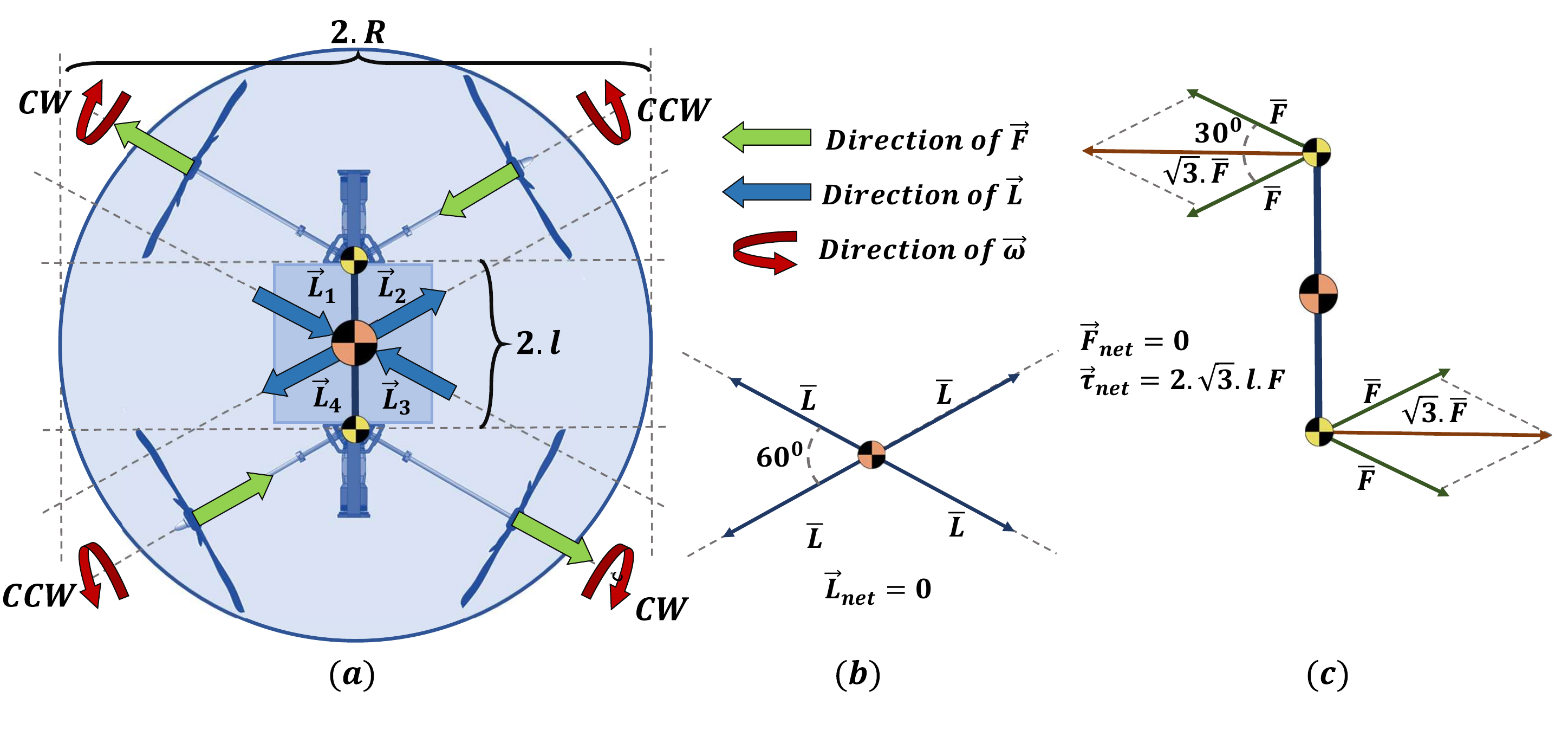}
  \caption{(a) Arrangement of skewed CW and CCW rotating propeller pairs in Spherical Shell of radius $R$, with the directions of produced thrust $\Vec{F}$, Angular momentum $\Vec{L}$, and the angular rotation $\Vec{\omega}$. (b) Neutralised net angular momentum $\Vec{L}_{net} = 0$. (c) Force Couple with net force $\Vec{F}_{net} = 0$ and torque $\Vec{\tau}_{net} = \sqrt{3}.l.F$.}\label{tnetlnet}
  \end{center}
\end{figure}

The propulsion mechanism of SPIROS is based on the principle of the Force Couple or the Pure Moment. According to the Varignon's Second Moment Theorem \cite{VaringnonTheorem}, the torque (moment)  of a couple is independent of the reference point. As shown in Fig.~\ref{tnetlnet}(a), a couple of two skewed clockwise (CW) and counterclockwise (CCW) rotating propellers are placed with $120^{0}$ angles between their axes. The direction of thrust produced by each propeller with equal magnitude $F$ is kept in such a way that the resultant angular momentum $\Vec{L}_{net}$ is neutralized (Fig.~\ref{tnetlnet}(b)), and Force Couple is formed with net torque (moment) $\Vec{\tau}_{net}= \sqrt{3}.l.F$ as shown in Fig.~\ref{tnetlnet}(c).

\begin{figure}[h]
  \begin{center}
  \includegraphics[width=\linewidth]{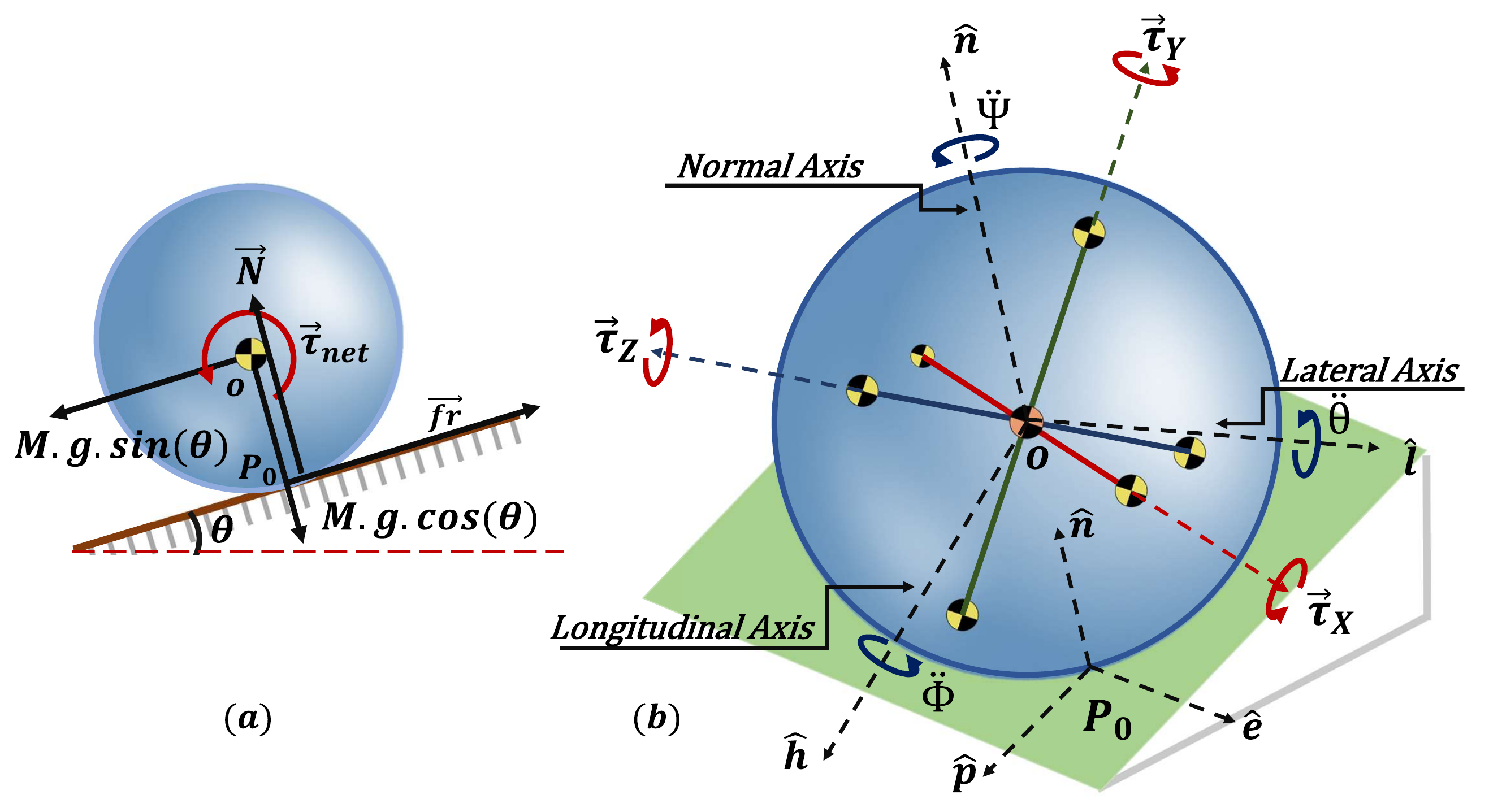}
  \caption{Schematic diagrams of (a) The basic forces and moments acting on the rolling Robot. (b) Reference frames and generalized robot dynamics parameters.}\label{wprinciple}
  \end{center}
\end{figure}

As shown in Fig.~\ref{wprinciple}(a), for the controlled maneuver over the slant surface, static friction $\Vec{fr}$ is utilized as an external force acting on the point of contact $P_{0}$. Given a surface slope $\theta$ relative to the gravity vector $\Vec{g}$, mass of the Robot $M$, the torque produced by the thrusters $\Vec{\tau}_{net}$, the angular acceleration of $\ddot{\zeta}$ can be expressed as:

\begin{equation}\label{zetaddot}
    \ddot{\zeta} = -\frac{\tau_{net}+ R.M.g.\sin(\theta)}{I_{0}+M.R^{2}}
\end{equation}

Whereas, $R$ is the radius of the sphere, and $I_{0}$ is the moment of inertia calculated about the axis of the rotation of the robot.

 Let, at an instant, the robot is rolling over the slant surface heading in $\hat{p}$ direction, $P_{0}$ is an instantaneous point of contact with the surface, $\hat{e}$ be the vector in the surface plane perpendicular to the direction of motion and $\hat{n}$ be the normal vector to the surface from $P_{0}$. Let the moments $\tau_{\hat{p}},\tau_{\hat{e}},\tau_{\hat{n}}$ are applied about axes $\hat{p}$, $\hat{e}$ and $\hat{n}$ to the robot the resultant angular acceleration vector $[\ddot{\zeta}_{\hat{p}},\ddot{\zeta}_{\hat{e}},\ddot{\zeta}_{\hat{n}}]^T $ can be obtained by substituting corresponding 3D quantities in eq. (\ref{zetaddot}), given by;

 \begin{equation}\label{zetaddot3D}
\begin{bmatrix}
\ddot{\zeta}_{\hat{p}}\\
\ddot{\zeta}_{\hat{e}}\\
\ddot{\zeta}_{\hat{n}}
\end{bmatrix}=-(\Tilde{I}_{0}+\Tilde{I}_{R})^{-1}\begin{bmatrix}
\tau_{\hat{p}}\\
\tau_{\hat{e}}-W_{||}R\\
\tau_{\hat{n}}
\end{bmatrix}
\end{equation}

Where, $W_{||}$ is the parallel component of the Weight vector $\Vec{W}$ along $\hat{p}$. The eq. (\ref{zetaddot3D}) is transformed in the terms of three moments $\tau_{\hat{X}}$,$\tau_{\hat{Y}}$ and $\tau_{\hat{Z}}$ along the three orthogonal axes of the robot frame $\mathscr{F}=\{\hat{X},\hat{Y},\hat{Z}\}$ (Fig.~\ref{tnetlnet}(c)). The instantaneous pose of the robot be $\alpha$ (yaw), $\beta$ (pitch) and $\gamma$ (roll). As shown in Fig.~\ref{wprinciple}(b), The resulting angular accelerations about arbitrary three mutually perpendicular axes of the frame $\mathscr{T}=\{\hat{h},\hat{l},\hat{n}\}$ are $\ddot{\Phi}$ along longitudinal axis $\hat{h}$, $\ddot{\Theta}$ along lateral axis $\hat{l}$, and $\ddot{\Psi}$ along normal axis $\hat{n}$ of the robot. The eq. (\ref{zetaddot3D}) is transformed to obtain angular acceleration vector $[\ddot{\Phi},\ddot{\Theta},\ddot{\Psi}]^{T}$ is given by:

\begin{equation}\label{genangddot}
    \begin{aligned}
    \begin{bmatrix}
     \ddot{\Theta}\\
     \ddot{\Phi}\\
     \ddot{\Psi}
    \end{bmatrix} = -R_{\xi\hat{n}}^{T}\Tilde{I}^{-1}R_{\hat{p}}R_{\alpha\beta\gamma}^{T}\begin{bmatrix}
     \tau_{\hat{X}}\\
     \tau_{\hat{Y}}\\
     \tau_{\hat{Z}}
    \end{bmatrix}-R_{\xi\hat{n}}^{T}\Tilde{I}^{-1}\begin{bmatrix}
     0\\
     W_{||}R\\
     0
    \end{bmatrix}
    \end{aligned}
\end{equation}

Where $R_{\alpha\beta\gamma}$ (eq. (\ref{Rabg})) is a general rotation matrix (Appendix \ref{'Rabc'}) gives rotational transform from robot frame $\mathscr{F}$ to the world frame $\mathscr{W}$. The matrices $R_{\xi\hat{n}}$ (eq. (\ref{Rxin})) and $R_{\hat{p}}$ (eq. (\ref{pen}) ) are the rotational transforms from the frame $\mathscr{T}$ to the frame $\{\hat{p},\hat{e},\hat{n}\}$ and the frame $\{\hat{p},\hat{e},\hat{n}\}$ to the world frame $\mathscr{W}$ respectively. (Appendix \ref{'ddot'}). $\Tilde{I} = \Tilde{I}_{0}+\Tilde{I}_{R})$ is the Inertia Tensor matrices calculated about $P_{0}$ illustrated in Appendix \ref{'Inot'}.

% ======================================================
% MECHANICAL DESIGN
% ======================================================

\section{Hardware Analysis}

To conceptualize the basic dynamic equation (\ref{genangddot}) of the robot with the required configurations of moments and thrusts discussed in Section II, the mechanical arrangement of the SPIROS is designed as follows.

\subsection{Intermeshing Rotary Thruster}

\begin{figure}[h]
  \begin{center}
  \includegraphics[width=0.95\linewidth]{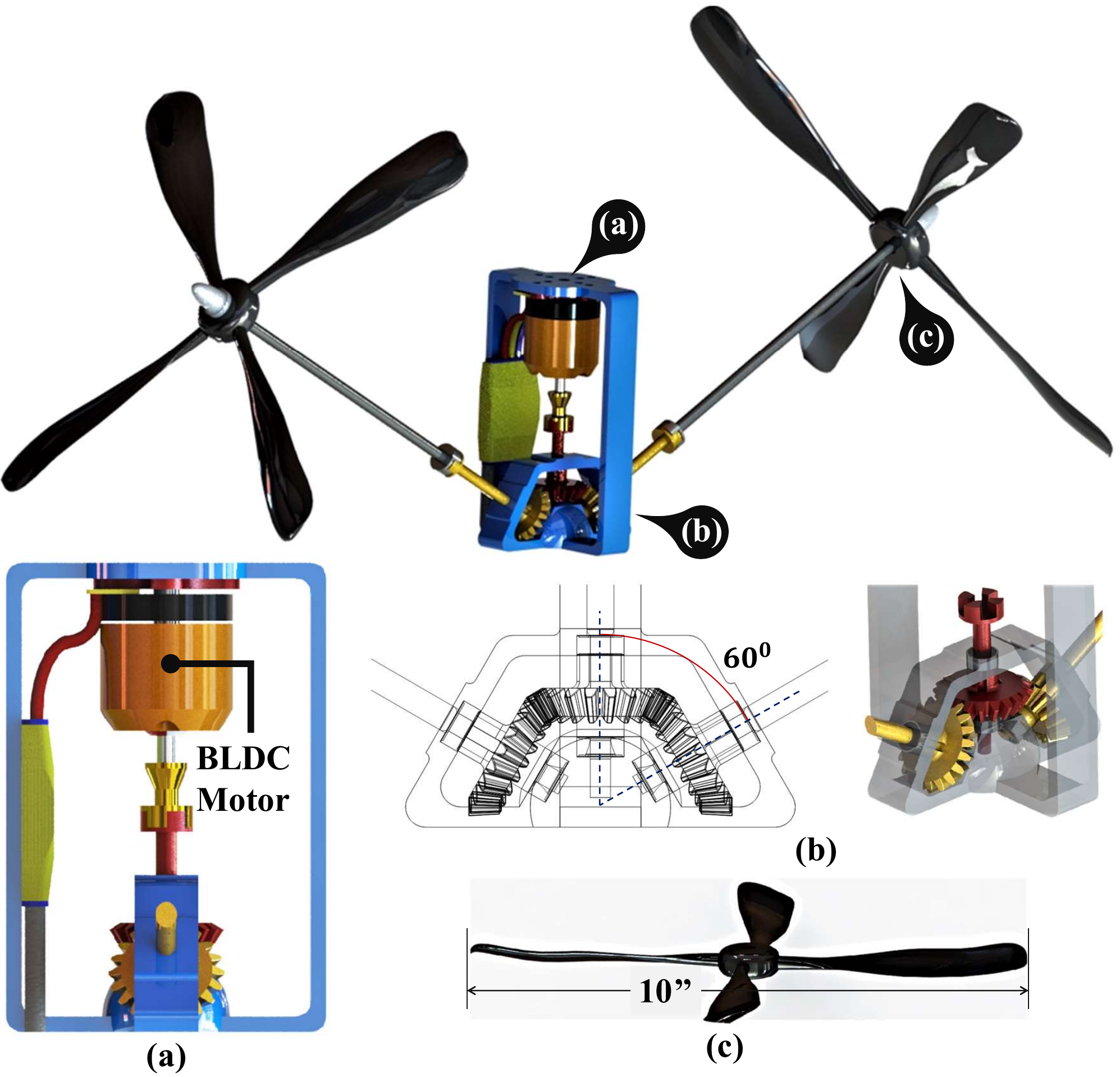}
  \caption{Schematic diagram of Intermeshing Rotary Thruster, (a) The BLDC motor with its Electronic Speed Controller (ESC). (b) The Bevel Gear mechanism. (c) 10" 4-blade Propeller}\label{Thruster}
  \end{center}
\end{figure}

The propeller thrust is the main driving force of the robot. The clockwise and anticlockwise rotation of the propeller pair shown in Fig~\ref{tnetlnet}, is incorporated by using the bevel gear \cite{reed1908bevel} arrangement. Three bevel gears with equal gear ratios are required with an angle $60^0$ with each other to transfer the driving rotational motion into the two $120^0$ skewed anti-rotating shafts. The schematic of the gearbox and the axle support with 3mm bearings are shown in the Fig.~\ref{Thruster}(b). The required outer shell diameter is minimized by the skewed axis of the shafts.

The thruster is actuated with the high-speed motors such as BLDC motors with its suitable ESC (Electronic Speed Controller) \cite{xia2012permanent}. The mechanical mounts for the driving motor and its speed controller are as shown in the Fig.~\ref{Thruster}(a). To obtain the required quantity of air thrust, the 10" four blades propellers are used (Shown in Fig.~\ref{Thruster}(c)). The overall structure of the Intermeshing Rotary Thruster is   modularly designed (as shown in Fig.~\ref{Thruster}) to repetitively construct and arrange in six-thruster configuration shown in Fig.~\ref{OverallConstruction}(a).

\subsection{Inner Support Cube}

\begin{figure}[h]
  \begin{center}
  \includegraphics[width=0.5\linewidth]{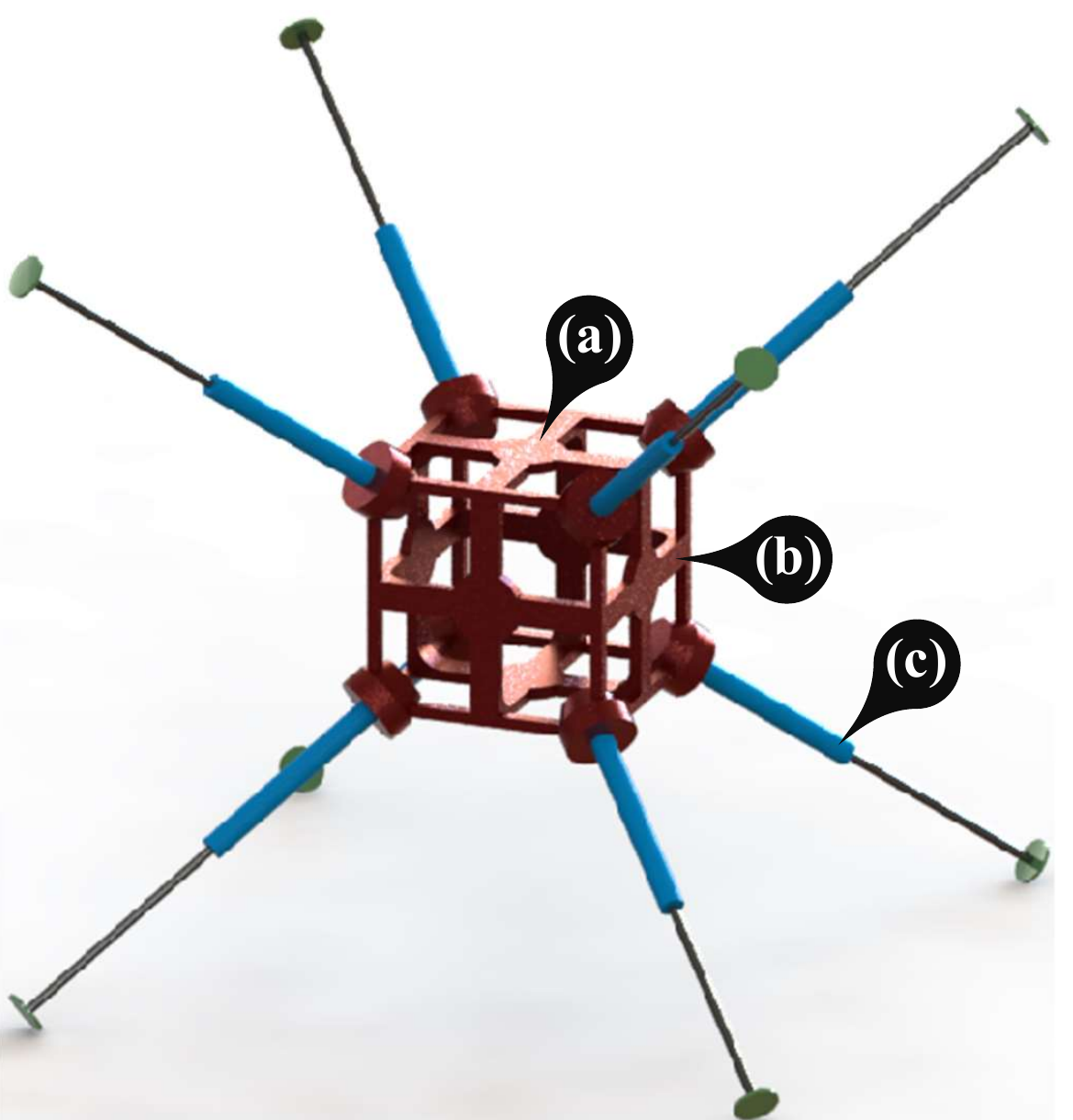}
  \caption{Schematic diagram of Inner Support Cube comprising of (a) The Thruster Mounts, (b) The Internal Hollow Frame, and (c) The Outer shell supports.}\label{InnerCube}
  \end{center}
\end{figure}

The inner support cube shown in Fig.~\ref{InnerCube} functions as the main supporting frame of the Robot's internal components. It comprises of thruster mounts, outer shell supports and a central hollow frame. The central hollow space provides room for the mechanical components as well as electronics other than thrusters.

Due to the central position of the structure, the Robot's non-inertial reference frame  $\mathscr{F}$ is defined along the three orthogonal vectors $\hat{X}$,$\hat{Y}$ and $\hat{Z}$ passing through center of the cube $O$ and centers of the respective faces.

The outer gridshell (Section III C.) is rigidly attached to the shell supports of the central cube. Due to the overall rolling motion of the robot, the internal framework, as well as whole thruster assembly, rolls with it.  Therefore, the transformation between the world frame $\mathscr{W}=\{\hat{x},\hat{y},\hat{z}\}$ and robot frame $\mathscr{F}$ is given by:

\begin{figure*}[!b]
  \begin{center}
  \includegraphics[width=\linewidth]{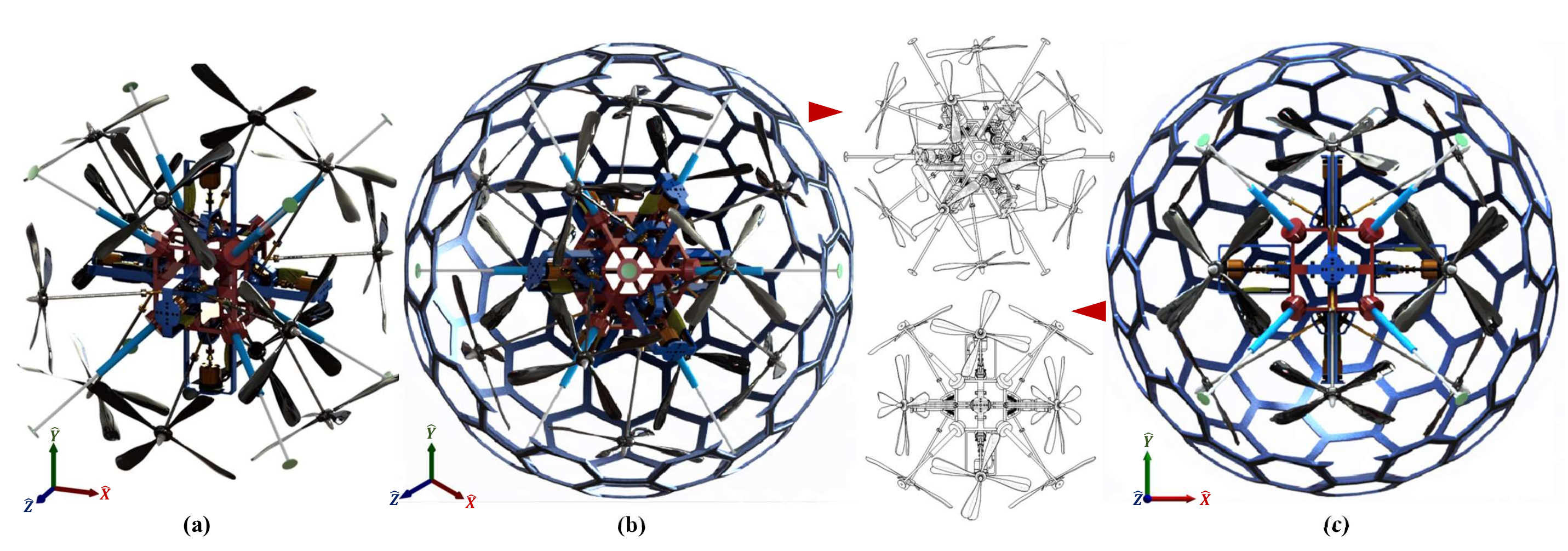}
  \caption{(a) Diametric View of the internal arrangement of the SPIROS, Schematics of the overall assemblies of the SPIROS in (b) Isometric View and (c) Front View.}\label{OverallConstruction}
  \end{center}
\end{figure*}

\begin{equation}
     \begin{bmatrix}
     \mathscr{W}_{3\times1}\\
     1
     \end{bmatrix}=\begin{bmatrix}
     R_{\alpha\beta\gamma} & P_{0}\\
     0&1
     \end{bmatrix}.\begin{bmatrix}
     \mathscr{F}_{3\times1}\\
     1
     \end{bmatrix}
 \end{equation}

\subsection{Outer Spherical Gridshell}

To ensure the free air circulation within the thrusters as well as to maintain structural rigidity and defined spherical geometry, the sphere-shaped perforated grid is used as an outer shell of the robot. Taking into account the structural properties of hexagon-pentagonal spherical shapes discussed in \cite{malek2013structural}\cite{martin1975ellipse}, such a grid shape offers rigidity and the free passage of the gases.

\begin{figure}[h]
  \begin{center}
  \includegraphics[width=\linewidth]{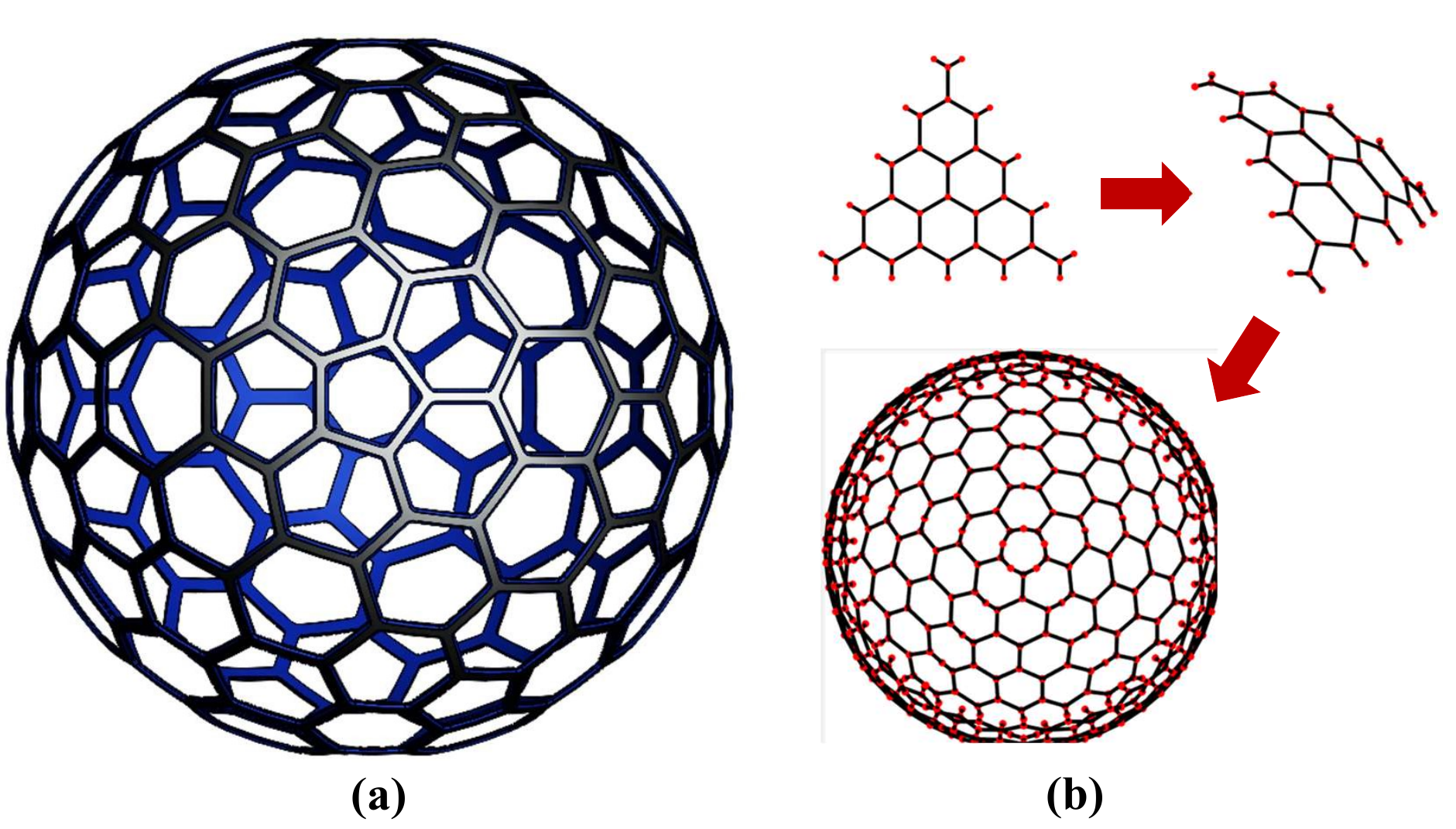}
  \caption{(a) The outer gridshell of the SPIROS, (b) Steps of building Goldberg Polyhedral grid.}\label{OuterShell}
  \end{center}
\end{figure}

The required shape of the grid considerably resembles the Goldberg polyhedron \cite{hart2013goldberg}. The method suggested in \cite{1937104} is applied to design the outer gridshell as illustrated in Fig.~\ref{OuterShell}. To get sufficient friction for pure rolling, the outer lining of the shell is coated with high friction coefficient materials as given in \cite{majidi2006high}.

\subsection{Electrical Architecture}

\begin{figure}[ht]
  \begin{center}
  \includegraphics[width=\linewidth]{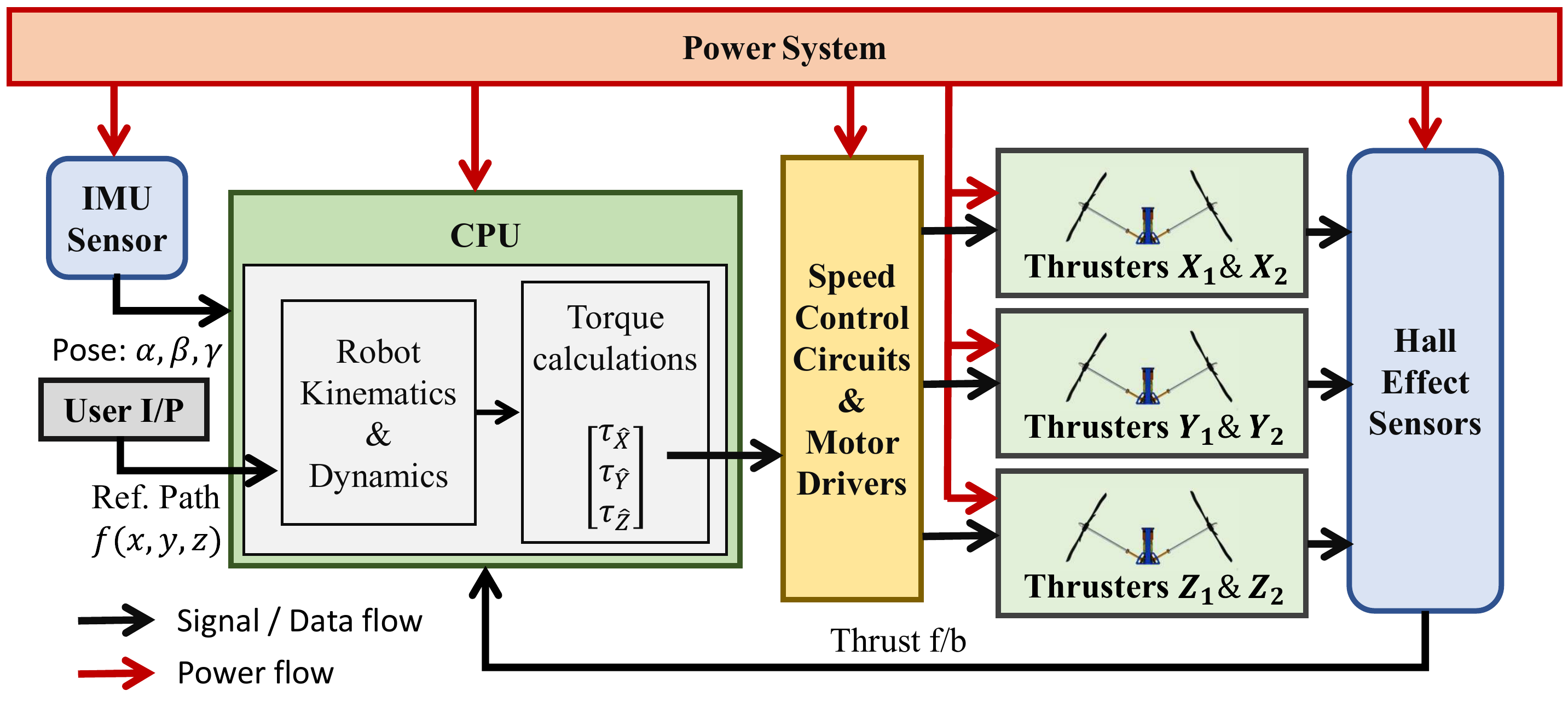}
  \caption{Electrical Architecture of the SPIROS.}\label{ElectArch}
  \end{center}
\end{figure}

The electrical architecture illustrated in Fig.~\ref{ElectArch} is utilized in the electrical system of the SPIROS. The reference path acquired from the user input is interpreted and the required torque vector is calculated. The angular speed signals are given to the thruster pairs via motor drivers. The IMU sensor \cite{chen1994gyroscope} and the Hall effect sensors are used to receive the robot's instantaneous pose $\{\alpha,\beta,\gamma\}$ and the output thrust magnitude's feedback during the course of motion.

\subsection{Overall Mechanical Assembly}

The thrusters (Section III A.) are attached to the central cubical frame (Section III B.) in the configuration schematically shown in Fig.~\ref{OverallConstruction}(a). To ensure the dynamic conditions proposed in the working principle (Section II.) of the robot, the magnitude of the thrust generated by the diametrically opposite thrusters are kept equal by the speed of the respective actuators. This arrangement enables the generation of the controlled moments $\tau_{\hat{X}}$, $\tau_{\hat{Y}}$ and $\tau_{\hat{Z}}$ along the axes of the robot frame $\mathscr{F}=\{\hat{X},\hat{Y},\hat{Z}\}$.  Table.~\ref{table:ThrustMoments} gives the geometric coordinates, thrust generated and the resultant torques by each thruster with respect to the robot frame $\mathscr{F}$.

\begin{table}[h!]
\centering
\caption{Geometric positions, the thrust generated and the resultant torques by each thruster with respect to the robot frame $\mathscr{F}$.}
\begin{tabular}{ c c c c }
\hline
No. & Position $\Vec{R}$ & Thrust $\Vec{F}$& Torque $\tau=\Vec{R}\times\Vec{F}$\\
\hline
\hline

1. & $[0,l,0]^{T}$ & $[0,0,F_{\hat{X}}]^{T}$ & $[lF_{\hat{X}},0,0]^{T}$\\
2. & $[0,-l,0]^{T}$ & $[0,0,-F_{\hat{X}}]^{T}$ & $[lF_{\hat{X}},0,0]^{T}$\\
3. & $[0,0,l]^{T}$ & $[F_{\hat{Y}},0,0]^{T}$ & $[0,lF_{\hat{Y}},0]^{T}$\\
4. & $[0,0,-l]^{T}$ & $[-F_{\hat{Y}},0,0]^{T}$ & $[0,lF_{\hat{Y}},0]^{T}$\\
5. & $[l,0,0]^{T}$ & $[0,F_{\hat{Z}},0]^{T}$ & $[0,0,lF_{\hat{Y}}]^{T}$\\
6. & $[-l,0,0]^{T}$ & $[0,-F_{\hat{Z}},0]^{T}$ & $[0,0,lF_{\hat{Y}}]^{T}$\\
\hline
\end{tabular}
\label{table:ThrustMoments}
\end{table}

%=======================================================
%  Kinematics and Dynamics Modelling.
%=======================================================

\section{Kinematics and Dynamics Modelling}

\begin{figure}[h!]
  \begin{center}
  \includegraphics[width=0.6\linewidth]{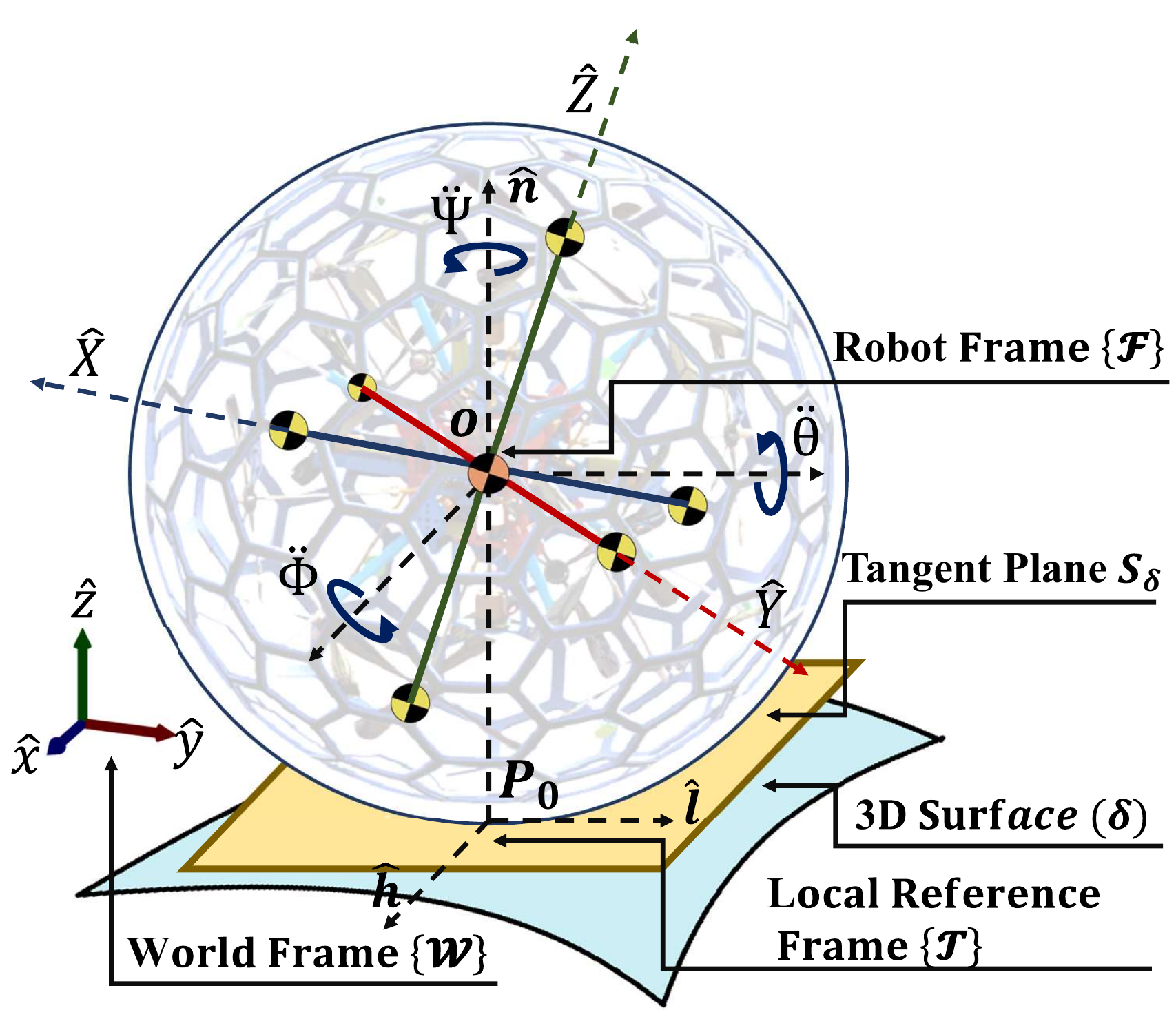}
  \caption{Schematic diagram of reference frames and points that are used to derive the kinematics and dynamics equation of the rolling SPIROS on a 3D surface $\sigma$.}\label{Dynamics}
  \end{center}
\end{figure}

Consider a spherical SPIROS robot rolling over a 3D surface $\sigma$ (Appendix \ref{'ddot'}), as shown in Fig.~\ref{Dynamics}. To derive the kinematics and the dynamics equations of rolling motion, the reference frames and the points are defined as follows. The position of the contact point of the robot with $\sigma$ is represented as $P_{0}$, which instantaneously indicates the position coordinates $[x(t),y(t),z(t)]^T$ of the robot in the inertial world frame $\mathscr{W}=\{\hat{x},\hat{y},\hat{z}\}$. The moments generated by the thruster assembly ($[\tau_{\hat{X}},\tau_{\hat{Y}},\tau_{\hat{Z}}]^T$) are considered along the axes of the non-inertial robot frame $\mathscr{F}=\{\hat{X},\hat{Y},\hat{Z}\}$. The robot frame rotates about the sphere center $O$ as the robot performs the rolling motion.

The frame $\mathscr{T}=\{\hat{h},\hat{l},\hat{n}\}$ is defined with its axes $\hat{h}$ and $\hat{l}$ laying on the imaginary tangent plane $S_{\sigma}$ passing through $P_{0}$. The normal axis $\hat{n}$ of the robot is defined perpendicular to the  $S_{\sigma}$ coinciding with the center of the sphere $O$, while lateral and longitudinal axes are defined parallel to the $S_{\sigma}$, That is, they do not rotate through rotation of the sphere about axes parallel to the $S_{\sigma}$, but they rotate accordingly when the sphere rotates about the normal axis $\hat{n}$. The rolling angle $\Theta$, tilting angle $\Phi$ and the turning angle $\Psi$ are as per defined in Section II.

Assumptions: The mathematical model of the SPIROS is derived considering the following assumptions:

\begin{enumerate}
    \item The surface $\sigma$ is rough enough to allow the sphere to roll over without slipping, skidding or falling; That is, without losing the traction with the surface.
    \item The surface $\sigma:z=f(x,y)$ is a continuous function of class $C_{2}$ or higher \cite{barsky1989geometric}.
    \item The outer gridshell is rigid, and the radius of the curvature of the surface $\sigma$ is never smaller than the radius of the gridshell, therefore, the robot remains in the single point of contact with the surface.
    \item The moments generated and the frictional force are always enough to provide the required driving force to the robot.
\end{enumerate}

Regarding to the kinematics equations derived by Moazami et al. \cite{moazami2019kinematics} based on the different categories of the SRs rolling over the 3D terrains, the transformation between the world frame $\mathscr{W}$ co-ordinates and the tangent plane frame $\mathscr{T}$ co-ordinates is given by:

\begin{equation}
    \begin{bmatrix}
    \mathscr{W}_{3\times1}\\
    1
    \end{bmatrix}=\begin{bmatrix}
    a_{11}&a_{12}&a_{13}&x\\
    a_{21}&a_{22}&a_{23}&y\\
    a_{31}&a_{32}&a_{33}&z\\
    0&0&0&1\\
    \end{bmatrix}\begin{bmatrix}
    \mathscr{T}_{3\times1}\\
    1
    \end{bmatrix}
\end{equation}

and,

\begin{equation}\label{hlnTrans}
    [\hat{h},\hat{l},\hat{n}]=\begin{bmatrix}
    a_{11}&a_{12}&a_{13}\\
    a_{21}&a_{22}&a_{23}\\
    a_{31}&a_{32}&a_{33}\\
    \end{bmatrix}
\end{equation}

where the elements $a_{jk}$ are given by:

\begin{equation}
\begin{aligned}
\begin{array}{l}
a_{11}=(1-s^{2}f_{x}^{2}(1-s_{n}))C\Psi+s^{2}f_{x}f_{y}(1-s_{n})S\Psi\\
a_{12}=(1-s^{2}f_{x}^{2}(1-s_{n}))S\Psi-s^{2}f_{x}f_{y}(1-s_{n})C\Psi\\
a_{13}=-s_{n}f_{x}\\
a_{21}=-s^{2}f_{x}f_{y}(1-s_{n})C\Psi-(1-s^{2}f_{y}^{2}(1-s_{n}))S\Psi \\
a_{22}=-s^{2}f_{x}f_{y}(1-s_{n})S\Psi+(1-s^{2} f_{y}^{2}(1-s_{n}))C\Psi\\
a_{23}=-s_{n}f_{y}\\
a_{31}=s_{n}f_{x}C\Psi-s_{n}f_{y}S\Psi\\
a_{32}=s_{n}f_{x}S\Psi+s_{n}f_{y}C\Psi\\
a_{33}=s_{n}\\
\end{array}
\end{aligned}
\end{equation}

while, $C\Psi$,~$S\Psi$,~$f_x$,~$f_y$,~$s$ and $s_{n}$ are the terms defined in Appendix \ref{'ddot'}.

The dynamic model of the SPIROS is derived considering the kinematic models of the three sub-categories of the Continuous Rolling Spherical Robot (CR-SR) Model borrowed from \cite{moazami2019kinematics}.

\subsection{Triple-Axis Rolling Spherical Robot (3R-SR) Model}

This is the most general category of the CR-SR. In this model, the SR is considered to rotate about all of its three axes, which is a case for the SPIROS. The kinematical model for the 3R-SR given in \cite{moazami2019kinematics} is as per the eq. (\ref{3RSRkinematics}).

\begin{equation}\label{3RSRkinematics}
    \begin{bmatrix}
    \Dot{x}\\
    \Dot{y}\\
    \Dot{z}
    \end{bmatrix}=R\mathscr{L}\begin{bmatrix}
    \Dot{\Theta}\\
    \Dot{\Phi}
    \end{bmatrix},\mathscr{L}=R\begin{bmatrix}
    a_{11}&-a_{12}\\
    a_{21}&-a_{22}\\
    a_{31}&-a_{32}
    \end{bmatrix}
\end{equation}

Therefore,

\begin{equation}\label{linv}
    \begin{bmatrix}
    \Dot{\Theta}\\
    \Dot{\Phi}
    \end{bmatrix}=\mathscr{L}^{\dagger}\begin{bmatrix}
    \Dot{x}\\
    \Dot{y}\\
    \Dot{z}
    \end{bmatrix},\mathscr{L}^{\dagger}=\frac{1}{R}\begin{bmatrix}
    \frac{a_{22}}{s_{n}^{2}}&-\frac{a_{12}}{s_{n}^{2}}&0\\
    0&-\frac{a_{31}}{f_{x}s_{n}}&\frac{a_{21}}{f_{x}s_{n}}
    \end{bmatrix}
\end{equation}

By differentiating eq.(\ref{3RSRkinematics}) on both sides and combining it with eq.(\ref{linv}) leads to:

\begin{equation}\label{3RSRdynamics}
   \begin{bmatrix}
    \Ddot{x}\\
    \Ddot{y}\\
    \Ddot{z}
    \end{bmatrix}=\Dot{\mathscr{L}}\mathscr{L}^{\dagger}\begin{bmatrix}
    \Dot{x}\\
    \Dot{y}\\
    \Dot{z}
    \end{bmatrix}+\begin{bmatrix}
    \mathscr{L}&0_{3\times1}
    \end{bmatrix}\begin{bmatrix}
    \Ddot{\Theta}\\
    \Ddot{\Phi}\\
    \Ddot{\Psi}
    \end{bmatrix}
\end{equation}

Therefore, from eq. (\ref{3RSRdynamics}) and eq. (\ref{genangddot}), the dynamics model is written in the following canonical form:

\begin{equation}\label{DynModel1}
    \Ddot{\mathscr{X}}_{\mathscr{W}}
    =\mathscr{A}_{3\times3}\Dot{\mathscr{X}}_{\mathscr{W}}+\mathscr{B}_{3\times3}\tau_{\mathscr{F}} + \mathscr{D}_{3\times1}
\end{equation}

Whereas,

\begin{equation}
    \begin{aligned}
        \begin{array}{l}
        \vspace{2mm}
            \mathscr{X}_{\mathscr{W}}=\begin{bmatrix}
            x\\
            y\\
            z
            \end{bmatrix},\tau_{\mathscr{F}}=\begin{bmatrix}
            \tau_{\hat{X}}\\
            \tau_{\hat{Y}}\\
            \tau_{\hat{Z}}
            \end{bmatrix}\\\vspace{5mm}
             \mathscr{A}_{3\times3} = -\Dot{\Psi}\begin{bmatrix}
             \frac{a_{12}a_{22}}{s_{n}^{2}}&-\frac{a_{11}a_{31}}{f_xs_{n}}-\frac{a_{12}^{2}}{s_{n}^{2}}&\frac{a_{11}a_{21}}{f_xs_{n}}\\
             \frac{a_{22}^{2}}{s_{n}^{2}}&-\frac{a_{21}a_{31}}{f_xs_{n}}-\frac{a_{12}a_{22}}{s_{n}^{2}}&\frac{a_{21}^{2}}{f_xs_{n}}\\
             \frac{a_{22}a_{32}}{s_{n}^{2}}&-\frac{a_{31}^{2}}{f_xs_{n}}-\frac{a_{21}a_{32}}{s_{n}^{2}}&\frac{a_{21}a_{31}}{f_xs_{n}}\\
             \end{bmatrix},\\\vspace{3mm}
             \mathscr{B}_{3\times3}=-\begin{bmatrix}
    \mathscr{L}&0_{3\times1}
    \end{bmatrix}R_{\xi\hat{n}}^{T}\Tilde{I}^{-1}R_{\hat{p}}R_{\alpha\beta\gamma}^{T},\\
             \mathscr{D}_{3\times1}=-\begin{bmatrix}
    \mathscr{L}&0_{3\times1}
    \end{bmatrix}R_{\xi\hat{n}}^{T}\Tilde{I}^{-1}\begin{bmatrix}
    0\\
    \frac{s_nMgR}{s}\\
    0
    \end{bmatrix}
        \end{array}
    \end{aligned}
\end{equation}

\subsection{Dual-Axis Rolling Spherical Robot (2R-SR) Model}

In this model, the robot is considered to be able to roll about its two main axis. As the rotation about the vertical axis is restricted, it is assumed that $\Dot{\Psi}=0$, therefore the eq. (\ref{DynModel1}) can be modified as:

\begin{equation}\label{DynModel2}
    \Ddot{\mathscr{X}}_{\mathscr{W}}
    =\mathscr{B}_{3\times3}\tau_{\mathscr{F}} + \mathscr{D}_{3\times1}
\end{equation}

Hence, the eq. (\ref{DynModel2}) represents the 2R-SR dynamic model of the SPIROS.

\subsection{Rolling and Turning Spherical Robot (RT-SR) Model}

In addition to the rolling action, RT-SRs are supposed to turn about their vertical axis to change their moving direction. The robot mechanism is assumed to provide angular velocities of $\Dot{\Theta}$ and $\Dot{\Psi}$ while $\Dot{\Phi}=0$. Therefore the eq. (\ref{3RSRkinematics}) and (\ref{linv}) can be rewritten as:

\begin{equation}\label{RTSRkinematics}
    \begin{bmatrix}
    \Dot{x}\\
    \Dot{y}\\
    \Dot{z}
    \end{bmatrix}=R\begin{bmatrix}
    a_{11}\\
    a_{21}\\
    a_{31}
    \end{bmatrix}\Dot{\Theta},~\Dot{\Theta}=[a_{11},a_{21},a_{31}]\begin{bmatrix}
    \Dot{x}\\
    \Dot{y}\\
    \Dot{z}
    \end{bmatrix}
\end{equation}

Hence, by differentiating eq. (\ref{RTSRkinematics}) we get;

\begin{equation}\label{RTSTdynamics}
 \begin{bmatrix}
    \Ddot{x}\\
    \Ddot{y}\\
    \Ddot{z}
    \end{bmatrix}=-\Dot{\Psi}\begin{bmatrix}
    a_{12}\\
    a_{22}\\
    a_{32}
    \end{bmatrix}[a_{11},a_{21},a_{31}]\begin{bmatrix}
    \Dot{x}\\
    \Dot{y}\\
    \Dot{z}
    \end{bmatrix}+R\begin{bmatrix}
    a_{11}\\
    a_{21}\\
    a_{31}
    \end{bmatrix}\Ddot{\Theta}
\end{equation}

Therefore, the matrices $\mathscr{A}$, $\mathscr{B}$ and $\mathscr{D}$ of the canonical form (eq. (\ref{DynModel1})) are modified as given below.

\begin{equation}\label{dyn3par}
    \begin{aligned}
        \begin{array}{l}
        \vspace{3mm}
             \mathscr{A}_{3\times3} = -\Dot{\Psi}\begin{bmatrix}
             a_{12}a_{12}&a_{12}a_{21}&a_{12}a_{31}\\
             a_{11}a_{22}&a_{21}a_{22}&a_{22}a_{31}\\
             a_{12}a_{31}&a_{22}a_{31}&a_{31}a_{32}\\
             \end{bmatrix},\\\vspace{3mm}
             \mathscr{B}_{3\times3}=-R.\begin{bmatrix}
    a_{11}&0&0\\
    a_{21}&0&0\\
    a_{31}&0&0
    \end{bmatrix}R_{\xi\hat{n}}^{T}\Tilde{I}^{-1}R_{\hat{p}}R_{\alpha\beta\gamma}^{T},\\
             \mathscr{D}_{3\times1}=-R.\begin{bmatrix}
    a_{11}&0&0\\
    a_{21}&0&0\\
    a_{31}&0&0
    \end{bmatrix}R_{\xi\hat{n}}^{T}\Tilde{I}^{-1}\begin{bmatrix}
    0\\
    \frac{s_nMgR}{s}\\
    0
    \end{bmatrix}
        \end{array}
    \end{aligned}
\end{equation}

%=======================================================
%  Path Tracking Control
%=======================================================

\section{Path Tracking Control}

In this section, the pure pursuit method is utilized to design the path tracking controls for the SPIROS based on the three dynamics models derived in Section IV.

Considering $\mathscr{P}=[x_d(t),y_d(t),z_d(t)]^T$ as a desired trajectory laying on $\sigma$, which is assumed to be continuous; that is,$\parallel[\Ddot{x}_d,\Ddot{y}_d,\Ddot{z}_d]^T\parallel$ is bounded for $t>0$. The stable error term $e_t$ with the deviation angle $\zeta$ between the heading direction $\hat{h}$ is defined as:

\begin{equation}
    e_t=[x_d(t),y_d(t),z_d(t)]^T-P_0
\end{equation}

The objective of this algorithm is to converge the
tracking error $e_t$ along with $\zeta$ to zero. The given path-planning expressions computes values for angular accelerations $\Ddot{\Theta}$, $\Ddot{\Phi}$ and $\Ddot{\Psi}$. Further, the eq. (\ref{genangddot}) is used to calculate the respective torques $\tau_{\hat{X}}$,$\tau_{\hat{Y}}$ and $\tau_{\hat{Z}}$.

\subsection{3R-SR Mode Path Tracking Control}

As 3R-SRs dynamic model is capable of providing $\Ddot{\Theta}$ $\Ddot{\Phi}$ and $\Ddot{\Psi}$, the controller is designed to provide the required values accordingly as follows:

\begin{equation}
\begin{array}{l}
\ddot{\Theta}=k_{\Theta} \frac{\|e_{t}\|}{(k_{e}+\|e_{t}\|)} \cos (\zeta)+\frac{1}{R}\|[\ddot{x}_{d}, \ddot{y}_{d}, \ddot{z}_{d}]^{T}\| \\
\ddot{\Phi}=-(k_{\Phi} \frac{\|e_{t}\|}{(k_{e}+\|e_{t}\|)} \sin (\zeta)+k_{\Phi_{2}} \sin (\zeta)) \\
\ddot{\Psi}=k_{\Psi} \zeta
\end{array}
\end{equation}

\subsection{2R-SR Mode Path Tracking Control}

As in 2RSR model, the rotation about the vertical axis is restricted, it can be implied that the robot can only use $\Ddot{\Theta}$ and $\Ddot{\Phi}$ to move towards the target. Therefore, the controller is given by:

\begin{equation}
\begin{array}{l}
\ddot{\Theta}=k_{\Theta_{1}} \frac{\|e_{t}\|}{(k_{e}+\|e_{t}\|)} \cos (\zeta)+k_{\Theta_{2}} \cos (\zeta) \\
\ddot{\Phi}=-(k_{\Phi_{1}} \frac{\|e_{t}\|}{(k_{e}+\|e_{t}\|)} \sin (\zeta)+k_{\Phi_{2}} \sin (\zeta)) \\
\ddot{\Psi}=0
\end{array}
\end{equation}

\begin{figure*}[hb]
  \begin{center}
  \includegraphics[width=0.8\linewidth]{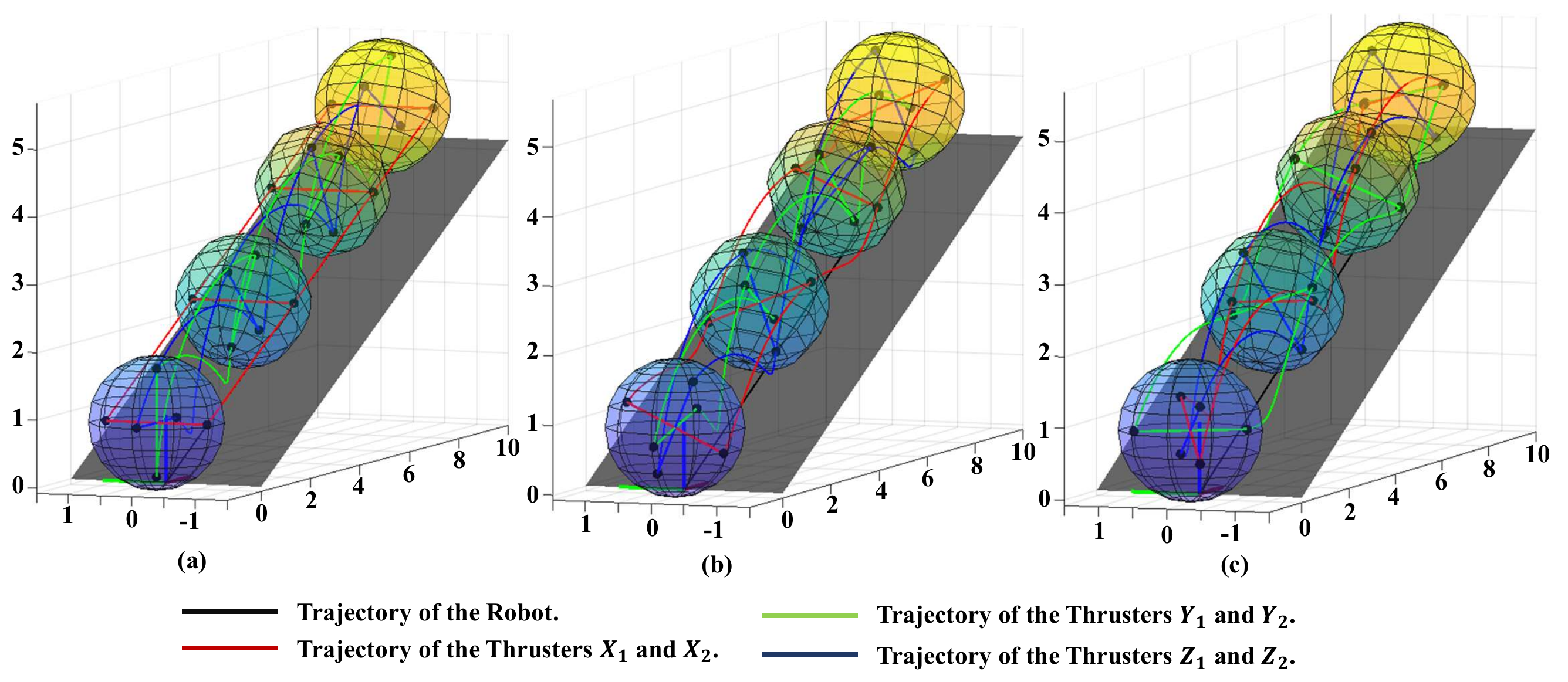}
  \caption{Robot ascending over the ramp with given initial poses (a)$\{\alpha=0^r,\beta=0^r,\gamma=0^r\}$, (b) $\{\alpha=\frac{\pi}{3}^r,\beta=\frac{\pi}{3}^r,\gamma=\frac{\pi}{3}^r\}$ and (c)$\{\alpha=\frac{\pi}{2}^r,\beta=\frac{\pi}{3}^r,\gamma=\frac{\pi}{6}^r\}$}\label{RampSimulation}
  \includegraphics[width=0.85\linewidth]{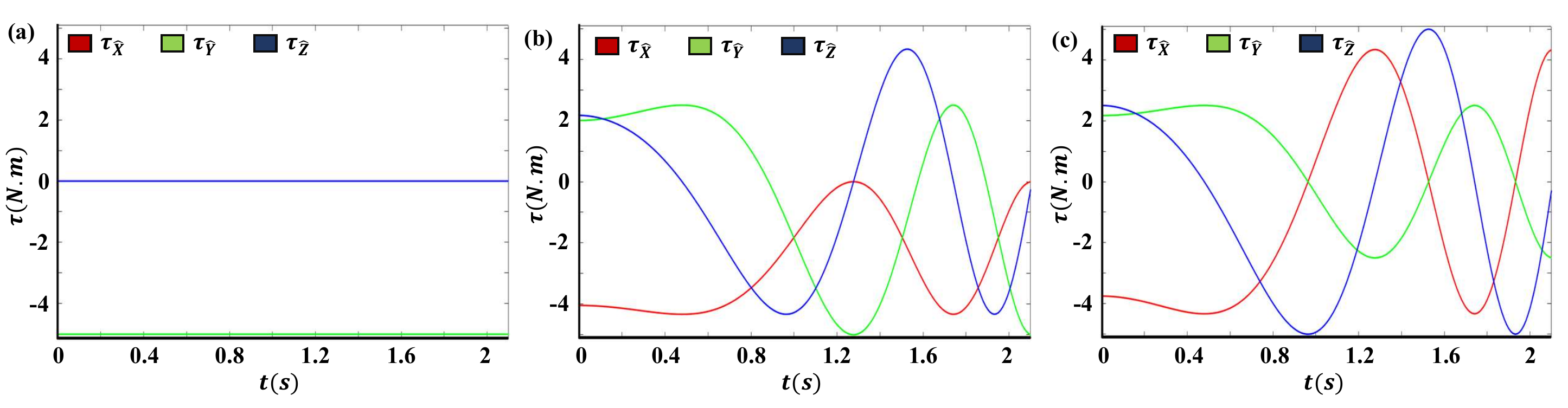}
  \caption{Robot's generated moments $\tau_{\hat{X}}$, $\tau_{\hat{Y}}$ and $\tau_{\hat{Z}}$ along the axes $\hat{X}$,$\hat{Y}$ and $\hat{Z}$ of the robot frame $\mathscr{F}$.}\label{RamptorqePlots}
  \end{center}
\end{figure*}

\subsection{RT-SR Mode Path Tracking Control}

As per the RT-SR model the robot can move towards the target point by compensating their deviation error $\zeta$, through turning action and then using $\ddot{\Theta}$ to approach and follow the target. This method is used to design the controller to provide the required angular accelerations as the following:

\begin{equation}
\begin{array}{l}
\ddot{\Theta}=k_{\Theta} \frac{\|e_{t}\|}{(k_{e}+\|e_{t}\|)} \cos(\zeta)+\frac{1}{R}\|[\ddot{x}_{d}, \ddot{y}_{d}, \ddot{z}_{d}]^{T}\|\\
\ddot{\Phi}=0\\
\ddot{\Psi}=k_{\Psi} \zeta
\end{array}
\end{equation}

%=======================================================
%  SIMULATION RESULTS
%=======================================================

\section{Simulation Results}

The SPIROS's mathematical model has been simulated in MATLAB and Simulink to verify the dynamics and the controllers presented in Sections IV and V. The physical parameters of the robot are normalized as radius $R = 1m$ and mass $M = 0.5 kg$ (considered as a homogenous sphere) for the simulations. Two operations of the SPIROS are been simulated for the analysis are as follows.

The robot is raised onto the rough ramp (providing enough frictional force for pure rolling) having a slope $\frac{\pi}{8}^{r}$ through the moments generated by thrusters. The experiment is repeated with the different initial poses $\{\alpha,\beta,\gamma\}$ of the robot to analyze the effect on the required torques. The time-lapsed positions of the simulated robot along with its trajectories with initial poses $\{\alpha=0^r,\beta=0^r,\gamma=0^r\}$, $\{\alpha=\frac{\pi}{3}^r,\beta=\frac{\pi}{3}^r,\gamma=\frac{\pi}{3}^r\}$ and $\{\alpha=\frac{\pi}{2}^r,\beta=\frac{\pi}{3}^r,\gamma=\frac{\pi}{6}^r\}$ are shown in Fig.~\ref{RampSimulation} (a),(b) and (c) respectively. Whereas, the corresponding torque magnitudes $\tau_{\hat{X}}$, $\tau_{\hat{Y}}$ and $\tau_{\hat{Z}}$ taken positive along axes $\hat{X}$,$\hat{Y}$, $\hat{Z}$ of robot frame $\mathscr{F}$ are given in the Fig.~\ref{RamptorqePlots} (a),(b) and (c).

Further, the robot is traversed over the continuous rough 3D surface $\sigma$ on the targeted trajectory obtained by projecting a circle over $\sigma$ such that:

\begin{equation}\label{trajectory}
\begin{array}{l}
    x=5+\sin\left(\frac{2\pi}{10}t\right)\\
    y=5+\cos\left(\frac{2\pi}{10}t\right)  
\end{array}
\end{equation}

Where the terrain surface, $\sigma$ is defined as the following: 

\begin{equation}\label{sigma}
    z=0.5\left(\cos\left(\frac{2\pi}{8}x\right)+\cos\left(\frac{2\pi}{8}y\right)-2\right)
\end{equation}

\begin{figure*}[ht!]
  \begin{center}
  \includegraphics[width=0.85\linewidth]{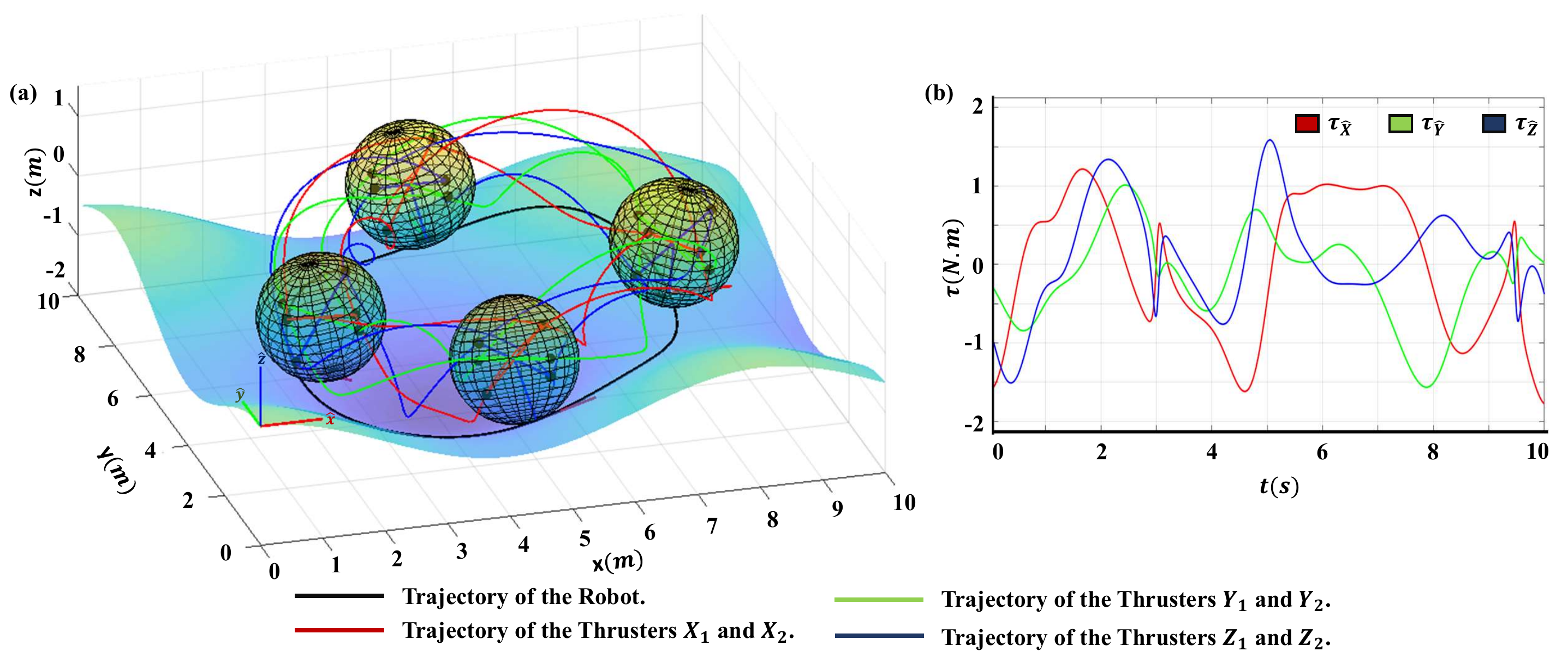}
  \caption{(a) Robot traversing over a 3D surface sigma in a smooth circular path. (b) Applied torques $\tau_{\hat{X}}$,$\tau_{\hat{Y}}$,$\tau_{\hat{Z}}$ by the thrusters. }\label{Circle_Simulation}
  \end{center}
\end{figure*}

With controller design parameters illustrated in Section V are tuned by the trial-and-error method, the robot successfully traced the circular trajectory (eq.\ref{trajectory}) shown by 3D time-lapse and trajectory plots in Fig.~\ref{Circle_Simulation}(a). Also, the magnitudes of the torque applied to the robot during the operation are plotted versus time in Fig.~\ref{Circle_Simulation}(b).  

%=======================================================
%    Conclusion and futureworks
%=======================================================

\section{Conclusion and Futureworks}

In this paper, the proposed SPIROS-Robot mechanism incorporating a novel six thruster propulsion system provides several advantages over the existing SRs. The working principle of the robot is explained. Then the design details of the mechanical structure and the electrical system discussed in detail.  By utilizing the present kinematics models for the SRs, the dynamic equations are derived along with the control schemes based on pure pursuit algorithm are given for the SPIROS. Finally, the derived dynamic equations of the SPIROS are mathematically modeled and numerically simulated in MATLAB and Simulink, showing the expected results.

The future work involves including the effects of the outer shell hindrance and the propeller interaction on the thrusts generated in the mathematical model by the advanced airflow analysis. Also improving on the control scheme to design a robust closed-loop controller. The next step is to develop a hardware prototype of the SPIROS having all the functionalities proposed in this paper. 


% ============================================

%Appendix one text goes here %\cite{Roberg2010}.

%\section{}

\appendices

\section{General Rotations}
\label{'Rabc'}

The following matrix $R_{\alpha\beta\gamma}$ represents a rotation whose yaw, pitch, and roll angles are $\alpha$, $\beta$ and $\gamma$ respectively.

\begin{equation}\label{Rabg}
    R_{\alpha\beta\gamma}=\begin{bmatrix}
      C\alpha C\beta & C\alpha C\beta S\gamma-S\alpha C\gamma & C\alpha S\beta C\gamma+S\alpha S\gamma \\
        S\alpha C\beta & S\alpha S\beta S\gamma+C\alpha C\gamma & S\alpha S\beta C\gamma-C\alpha S\gamma \\
      -S\beta & C\beta S\gamma & C\beta C\gamma
    \end{bmatrix}
\end{equation}

Where $S\theta$ and $C\theta$ depicts $\sin(\theta)$ and $\cos(\theta)$ respectively. The pre-multiplication of $R_{\alpha\beta\gamma}$ with the 3D vector will produce the intrinsic rotated vector whose Tait–Bryan angles are $\alpha$, $\beta$ and $\gamma$ about axes $\hat{z}$, $\hat{y}$, $\hat{x}$, respectively.

\section{}
\label{'ddot'}

According to Fig.~\ref{wprinciple}, $\Theta$ is the sphere's rolling angle, that is, the rotation angle of the sphere about its lateral axis. $\Phi$ is the sphere's tilting angle, which is, the rotation angle of the sphere about its longitudinal axis, and $\Psi$ is the sphere's turning angle about the robot-side normal axis. Let $\sigma$ be a $3\mathrm{D}$ surface defined as $\sigma: F(x, y, z)=0$ in the form of $f(x, y)-z=0$ implying that for any $x$ and $y$, the elevation of the surface $\sigma$ along the vertical axis of $\hat{z}$ is $z=f(x, y)$.

The normal vector of surface $\sigma$ at position $P_{0}$ is given by

\begin{equation}\label{n}
\hat{n}=-\frac{\nabla f}{\parallel\nabla f \parallel}=s_{n}[-f_{x},-f_{y},1]^{T}
\end{equation}

where $f_{x}=\frac{\partial f(x, y)}{\partial x}$ and $f_{y}=\frac{\partial f(x, y)}{\partial y}$ are the partial derivative functions and $s_{n}=(f^{2}_{x}+f^{2}_{y}+1)^{-\frac{1}{2}}$. As $\hat{e}$ is perpendicular to the direction of motion and the normal $\hat{n}$.

\begin{equation}\label{e}
  \hat{e}=\hat{n}\times\hat{z} = s[f_{y},-f_{x},0]^{T}
\end{equation}

whereas, $s=(f^{2}_{x}+f^{2}_{y})^{-\frac{1}{2}}$. The vector $\hat{p}$ is along the direction of motion, hence it is perpendicular to the $\hat{e}$ and $\hat{n}$.

\begin{equation}\label{p}
  \hat{p}=\hat{e}\times\hat{n}=ss_{n}[-f_{x},-f_{y},-f^{2}_{x}-f^{2}_{y}]^{T}
\end{equation}

therefore, defining:

\begin{equation}\label{pen}
  R_{\hat{P}}=[\hat{p},\hat{e},\hat{n}]=\begin{bmatrix}
  -ss_{n}f_{x} & sf_{y} & -s_{n}f_{x}\\
  -ss_{n}f_{y} & -sf_{x} & -s_{n}f_{y}\\
  -ss_{n}(f^{2}_{x}+f^{2}_{y}) & 0 & s_{n}
  \end{bmatrix}
\end{equation}

According to the eq. (\ref{hlnTrans}), the vector $\hat{l}$ is given by:

\begin{equation}\label{l}
  \hat{l}=\begin{bmatrix}
  (1-s^{2}f^{2}_{x}(1-s_{n}))S\Psi-s^{2}f_{x}f_{y}(1-s_{n})C\Psi\\
  (1-s^{2}f^{2}_{x}(1-s_{n}))C\Psi-s^{2}f_{x}f_{y}(1-s_{n})S\Psi\\
  s_{n}f_{x}S\Psi+s_{n}f_{y}C\Psi
  \end{bmatrix}
\end{equation}

As vectors $\hat{p}$,$\hat{e}$,$\hat{h}$ and $\hat{l}$ are co-planar as they hold common normal vector $\hat{n}$, let acute angle between $\hat{p}\perp\hat{e}$ and $\hat{h}\perp\hat{l}$ be $\xi$. Therefore,

\begin{equation}\label{xi}
\begin{array}{l}
     \cos(\xi)=\hat{l}.\hat{e}= sf_{y}sin\Psi-sf_{x}\cos\Psi,\\
     \sin(\xi)=|\hat{l}\times\hat{e}|= sf_{x}\sin\Psi+sf_{y}\cos\Psi
\end{array}
\end{equation}

Now, from the eq. (\ref{zetaddot3D}), the $\{\hat{p},\hat{e},\hat{n}\}$ frame with the angular accelerations $\ddot{\zeta}_{\hat{p}}$,$\ddot{\zeta}_{\hat{e}}$ and $\ddot{\zeta}_{\hat{n}}$ are produced by moments $\tau_{\hat{p}}$, $\tau_{\hat{e}}$ and $\tau_{\hat{n}}$, leads to:

\begin{equation}\label{penzddot}
\begin{bmatrix}
\ddot{\zeta}_{\hat{p}}\\
\ddot{\zeta}_{\hat{e}}\\
\ddot{\zeta}_{\hat{n}}
\end{bmatrix}=-(\Tilde{I}_{0}+\Tilde{I}_{R})^{-1}\begin{bmatrix}
\tau_{\hat{p}}\\
\tau_{\hat{e}}\\
\tau_{\hat{n}}
\end{bmatrix}-(\Tilde{I}_{0}+\Tilde{I}_{R})^{-1}\begin{bmatrix}
0\\
W_{||}R\\
0
\end{bmatrix}
\end{equation}

 Defining $\Tilde{I}$ as an overall inertia tensor matrix being $\Tilde{I} = \Tilde{I}_{0}+\Tilde{I}_{R}$. Where the $\Tilde{I}_{0}$ is the inertia tensor matrix calculated about the center of gravity (CG) and $\Tilde{I}_{R}$ is the inertia tensor matrix calculated about point of contact $P_{0}$ considering the robot as a point mass concentrated at CG. $W_{||}$ is the parallel component of the Weight vector $\Vec{W}$ along $\hat{p}$.

 \begin{equation}
     W_{||} = [0,0,-Mg]^{T}.\hat{p}=ss_{n}(f^{2}_{x}+f^{2}_{y})Mg
 \end{equation}

 Considering (\ref{xi}) the transform between $[\ddot{\zeta}_{\hat{p}},\ddot{\zeta}_{\hat{e}},\ddot{\zeta}_{\hat{n}}]^{T}$ and $[\ddot{\Theta},\ddot{\Phi},\ddot{\Psi}]^{T}$ is given by,

 \begin{equation}\label{Rxin}
  \begin{array}{l}
      [\ddot{\zeta}_{\hat{p}},\ddot{\zeta}_{\hat{e}},\ddot{\zeta}_{\hat{n}}]^{T} = R_{\xi\hat{n}}[\ddot{\Theta},\ddot{\Phi},\ddot{\Psi}]^{T},\\
      R_{\xi\hat{n}}=\begin{bmatrix}
      -sf_{x}S\Psi-sf_{y}C\Psi&-sf_{x}C\Psi+sf_{y}S\Psi&0\\
      -sf_{x}C\Psi+sf_{y}S\Psi&sf_{x}S\Psi+sf_{y}C\Psi&0\\
      0&0&1
      \end{bmatrix}
  \end{array}
 \end{equation}

Similarly, the relation between the moments $[\tau_{\hat{p}},\tau_{\hat{e}},\tau_{\hat{n}}]^{T}$ and $[\tau_{\hat{X}},\tau_{\hat{Y}},\tau_{\hat{Z}}]^{T}$ is given by,

\begin{equation}\label{TxyzTpen}
[\tau_{\hat{p}},\tau_{\hat{e}},\tau_{\hat{n}}]^{T}=R_{\hat{p}}R_{\alpha\beta\gamma}^{T}[\tau_{\hat{X}},\tau_{\hat{Y}},\tau_{\hat{Z}}]^{T},R_{\hat{p}}=[\hat{p},\hat{e},\hat{n}]
\end{equation}

Therefore, by combining the eq. (\ref{penzddot}),eq. (\ref{Rxin}) and eq. (\ref{TxyzTpen}) leads to the eq. (\ref{genangddot}).

\section{Inertia Tensor $\Tilde{I}$}
\label{'Inot'}

Consider a 3D object having mass $M$, undergoing general motion in the frame of reference $\mathscr{W}=\{\hat{x},\hat{y},\hat{z}\}$, the inertia tensor $\Tilde{I}$ of the object with respect to $\mathscr{W}$ is defined by:

\begin{equation}\label{inertiatensor}
\Tilde{I}=\begin{bmatrix}
\int(y^2+z^2).dM&-\int xy.dM&-\int zx.dM\\
-\int xy.dM&\int(z^2+x^2).dM&-\int yz.dM\\
-\int zx.dM&-\int yz.dM&\int(x^2+y^2).dM
\end{bmatrix}
\end{equation}

$\Tilde{I}$ depends on the geometry of the object. The matrix $\Tilde{I}$ is constructed with respect to specified origin and orthogonal axis. According to the parallel axis theorem, if inertia tensor $\Tilde{I}_{cm}$ is known about the CG of the object then $\Tilde{I}$ about any parallel axis is given by $\Tilde{I}=\Tilde{I}_{cm}+\Tilde{I}_{R}$ whereas $\Tilde{I}_{R}$ matrix is obtained by treating the object as a point mass at the CG.

\section*{Acknowledgment}

This work was supported by IvLabs, Visvesvaraya National Institute of Technology, Nagpur, India.

\ifCLASSOPTIONcaptionsoff
  \newpage
\fi



%\begin{thebibliography}{1}

\bibliographystyle{IEEEtran}
\bibliography{IEEEabrv,Bibliography}
%\end{thebibliography}


% \newpage

% ==== SWITCH OFF the BIO for submission
% ==== SWITCH OFF the BIO for submission


\vfill

\end{document}